\documentclass[acmlarge, nonacm, pdftex]{acmart}

\setcopyright{rightsretained}
\acmJournal{IMWUT}
\acmYear{2024} \acmVolume{8} \acmNumber{2} \acmArticle{56}
\acmMonth{6}\acmDOI{10.1145/3659604}

\usepackage[utf8]{inputenc}
\usepackage{tcolorbox}
\usepackage{caption}
\usepackage{subcaption}
\usepackage[T1]{fontenc}
\usepackage{markdown}
\newcommand{\squishlist}{\begin{itemize}[itemsep=1pt,parsep=2pt,topsep=3pt,partopsep=0pt,leftmargin=0em, itemindent=1em,labelwidth=1em,labelsep=0.5em]}
\newcommand{\squishend}{\end{itemize}}

\newcommand\review[1]{\textcolor{black}{#1}}

\begin{document}

\title[From Classification to Clinical Insights]{From Classification to Clinical Insights: Towards Analyzing and Reasoning About Mobile and Behavioral Health Data With Large Language Models}

\author{Zachary Englhardt}
\authornote{Both authors contributed equally to this research.}
\email{zacharye@cs.washington.edu}
\orcid{0000-0002-6646-6466}
\orcid{1234-5678-9012}
\author{Chengqian Ma}
\authornotemark[1]
\orcid{0009-0002-3841-8966}
\affiliation{
  \institution{University of Washington}
  \city{Seattle}
  \state{Washington}
  \country{USA}
}

\author{Margaret E. Morris}
\orcid{0000-0002-8925-9718}

\author{Chun-Cheng Chang}
\orcid{0009-0000-5472-4897}
\affiliation{
  \institution{University of Washington}
  \city{Seattle}
  \state{Washington}
  \country{USA}
}

\author{Xuhai "Orson" Xu}
\orcid{0000-0001-5930-3899}
\affiliation{
  \institution{Massachusetts Institute of Technology}
  \city{Cambridge}
  \state{Massachusetts}
  \country{USA}
}

\author{Lianhui Qin}
\orcid{0009-0002-4757-8136}
\affiliation{
  \institution{University of California, San Diego}
  \city{La Jolla}
  \state{California}
  \country{USA}
}

\author{Daniel McDuff}
\orcid{0000-0001-7313-0082}

\author{Xin Liu}
\orcid{0000-0002-9279-5386}
\affiliation{
  \institution{University of Washington}
  \city{Seattle}
  \state{Washington}
  \country{USA}
}

\author{Shwetak Patel}
\orcid{0000-0002-6300-4389}

\author{Vikram Iyer}
\email{vsiyer@uw.edu}
\orcid{0000-0002-3025-7953}
\affiliation{
  \institution{University of Washington}
  \city{Seattle}
  \state{Washington}
  \country{USA}
}

\renewcommand{\shortauthors}{Englhardt and Ma, et al.}

\begin{abstract}
  Passively collected behavioral health data from ubiquitous sensors \review{could 
  provide mental health professionals valuable insights into patient's daily lives, but such efforts are impeded by disparate metrics, lack of interoperability, and unclear} 
  correlations between the measured signals and an individual's mental health. To address these challenges, we 
pioneer the exploration of large language models (LLMs) to synthesize clinically relevant insights from multi-sensor data. We develop chain-of-thought prompting methods to generate LLM reasoning \review{on how 
  data pertaining to activity, sleep and social interaction} 
  relate to conditions such as depression and anxiety. We then prompt the LLM to perform binary classification, achieving accuracies of 61.1\%, exceeding the 
state of the art. \review{We find models like GPT-4 correctly reference numerical data 75\% of the time.}

While we began our investigation by developing methods to use LLMs to output binary classifications for conditions like depression, we find instead that their greatest potential value to clinicians lies not in diagnostic classification, but rather in rigorous analysis of diverse self-tracking data to generate natural language summaries that synthesize multiple data streams and identify potential concerns. Clinicians envisioned using these insights in a variety of ways, principally for fostering collaborative investigation with patients to strengthen the therapeutic alliance and guide treatment. We describe this collaborative engagement, additional envisioned uses, and associated concerns that must be addressed before adoption in real-world contexts.

\end{abstract}






\maketitle

\section{introduction}
Mobile and wearable sensors that collect health and fitness data have seen explosive growth over the past five years~\cite{indrakumari2020growing, hickey2021rise}. The sensing capabilities of products such as Fitbit and Apple Watch have dramatically advanced beyond simple step counts to include optical heart rate sensors and two lead electrocardiogram (ECG) measurements that can provide clinicians valuable information about a patient's symptoms 
outside their practice. Beyond simply tracking a run, or checking for high heart rate,
researchers have shown the potential of leveraging such passive sensing data to model high-level, complex behaviors and mental health. \cite{xu_globem_2023,wang_crosscheck:_2016,wang_tracking_2018,salekin_weakly_2018}.

Despite their potential, mobile and wearable sensor data use in clinical mental health practice faces four key challenges. First, there is a lack of trust and clinically validated data; 
for instance, there is high uncertainty about \review{the specific relationships} between \review{mobile health signals (such as activity levels)} and mental states like depression \review{\cite{rohani_correlations_2018}}. Second, clinicians struggle with \review{the additional burden of incorporating mobile health data during treatment and often have} difficulty interpreting non-standard signals \review{generated by wearable devices and mobile phones~\cite{isakadze_how_2020, triantafyllidis_features_2019}. Third, conventional machine learning (ML) approaches for interpreting sensor data perform poorly on abstract relations. For example, recent work evaluating the ability of 19 different ML models to predict depression using self-tracking data revealed that many achieved accuracies below 50\%~\cite{xu_globem_2023}. 
Finally, the qualitative aspects of mental health further limit the utility of binary data classifications since diagnoses on their own contain no context about the patient's lifestyle and history, which are critical to designing treatment plans~\cite{chancellor_methods_2020}.}

\review{Given the challenges inherent in sensor- and ML-based approaches for health data analysis, it becomes imperative to explore alternative methods that offer}
flexibility, adaptability, and deeper analytical capabilities. This paper therefore considers the potential of LLMs to transform the way we approach sensor data in health settings. Unlike traditional ML techniques, LLMs can generate reasoning that incorporates cross-domain knowledge, potentially enabling the processing of multifaceted sensor data with greater contextual understanding~\cite{bubeck_sparks_2023}. This shift in methodology from rigid, classification-focused models to LLM's more versatile and interpretative capabilities offers a promising avenue for advancing health data analysis.

In this context, we contribute the following key insights and findings:


\begin{enumerate}
\item \textit{We perform the first exploration of LLM use to process multi-sensor ubiquitous wearable data.} We develop a series of new prompting and model fine tuning strategies that enable LLMs to perform zero-shot depression classification using raw, multi-modal wearable sensor data. We explore the effects of data input formats and numerical accuracy on state-of-the-art LLMs like GPT-4~\cite{openai2023gpt4}, PaLM 2~\cite{chowdhery2022palm} and GPT-3.5~\cite{qin2023chatgpt}. We demonstrate that these strategies outperform classical ML methods on state-of-the-art depression classification and highlight performance variations across models.
\item As an intermediate step to classification, \textit{we observe that LLMs can generate text reasoning about multi-sensor data,}  correctly describing trends and anomalies in the data and making connections between multiple input signals and relevant mental health scenarios.
\item To evaluate the feasibility of this approach, \textit{we undertake an interactive interview study with mental health professionals to critically assess the practicality and limitations of interpreting mobile health data using LLMs, }gain insight into the clinical relevance of LLM-generated reasoning, and to consider the prospective role of LLMs in mental health contexts. We find that clinicians express a strong desire to have access to LLM-based tools for collaborative investigation of mobile health data with patients, and \textit{we outline a series of scenarios for clinician-patient-AI collaborative therapy}.
\end{enumerate}

\section{related-work}
\label{sec:background}

\subsection{Multi-sensor Passive Sensing for Health and Well-being}
Smartphones and wearable devices, now ubiquitous in our lives, function as passive sensors, seamlessly capturing a vast range of data. Their near-constant presence enables unobtrusive and continuous monitoring of behavior, activity, and physiological signals. 
Over the last decade, significant progress has been made in passive sensing and behavioral modeling, impacting areas such as physiological health condition detection~\cite{10.1145/3090051,min2014_10.1145/2556288.2557220,xu_listen2cough_2021}, monitoring mental health status~\cite{wang2014studentlife,xu_leveraging_2021}, measuring job performance~\cite{10.1145/3328908,mattingly2019tesserae}, tracking education outcomes~\cite{wang2015smartgpa,zhang_impact_2022}, and tracing social justice~\cite{yasaman_2019_10.1145/3359216}. Researchers employ various methods, including statistical analysis and conventional ML models, to explore these areas. 

In the mental health context, initial research established statistical correlations between mental health conditions and mobile sensing data. For instance, \citet{saeb_mobile_2015} identified significant correlations between depression scores and smartphone usage patterns, and \citet{ben2015next} identified links between changes in depression severity levels and features related to sleep duration, speech duration, and mobility. More recent efforts have focused on leveraging these results to build ML models for mental health disorder diagnosis and detection~\cite{wahle_mobile_2016, chikersal_detecting_2021, nickels2021toward, wang_tracking_2018}. To further growth in this area, \citet{xu_globem_2023} collected and released a multi-year passive sensing dataset and platform that covers a wide range of physical health, mental health, and social well-being measurements. However, most research in this domain relies on conventional statistical and ML methods, and the recent improvements in the performance of large foundation models \cite{bubeck_sparks_2023} present an opportunity to explore new techniques for analyzing passively collected sensor data.

\subsection{LLMs for Health Applications}
The success of transformer-based language models, such as BERT~\cite{devlin_bert_2019} and GPT~\cite{radford_improving_2018}, has led to the development of larger and more powerful language models (e.g., GPT-3~\cite{brown_language_2020} and T5~\cite{raffel_exploring_2020}). Instruction fine-tuning by including instructions (i.e., prompts) from a range of datasets and task domains during both the training and generation phases has led to the development of single models that can perform a wide range of tasks~\cite{wei_finetuned_2022}. These instruction-fine-tuned LLMs, such as GPT-4~\cite{openai2023gpt4}, PaLM~\cite{chowdhery_palm_2022}, FLAN-T5~\cite{chung_scaling_2022}, LLaMA~\cite{touvron_llama_2023}, and Alpaca~\cite{taori_stanford_2023}, contain tens to hundreds of billions of parameters and achieve a promising level of performance on a variety of tasks, such as question answering~\cite{omar2023chatgpt,robinson2023leveraging}, logic reasoning~\cite{wei_chain--thought_2023,zhou_least--most_2023}, machine translation~\cite{brants2007large,gulcehre2017integrating}, and more.

In the health sector, these LLMs have been applied in several studies~\cite{jiang_health_2023,singhal_towards_2023,li_chatdoctor_2023,liu_large_2023,wu_pmc-llama_2023,nori_capabilities_2023}. For example, \citet{singhal_towards_2023} utilized a fine-tuned version of  PaLM-2 to score as high as 86.5\% on the MedQA dataset. Similarly, \citet{wu_pmc-llama_2023} fine-tuned LLaMA on a corpus of academic medical papers and textbooks, yielding  promising results on multiple biomedical QA datasets. \citet{jiang_health_2023} trained a medical language model on unstructured clinical notes from the electronic health record and fine-tuned for performance across a wide range of clinical and operational predictive tasks. These examples underscore the versatility and potential effectiveness of LLMs in the medical space. 

In the mental health domain, LLMs have been explored for applications such as sentiment analysis and emotional reasoning~\cite{qin2023chatgpt,zhong2023can,kocon2023chatgpt}. \citet{lamichhane_evaluation_2023}, \citet{amin_will_2023}.  
\citet{yang_evaluations_2023} tested the performance of ChatGPT on multiple classification tasks (stress, depression, and suicide risk) and found that it shows initial potential for these mental health applications, but it has room for significant improvement. 

Despite this past work, scant research focuses specifically on \textit{integration with mobile and wearable health data}, with most of the existing literature exploring text data rather than multi-sensor streams. Closer to our work, \citet{liu2023large} demonstrated that with only few-shot tuning, a LLM can ground various physiological and behavioral time-series data and make meaningful inferences on numerous health tasks (e.g., heart rate measurement, atrial fibrillation detection, and mood score prediction). However, their work is based on self-curated toy datasets consisting of well-described physiological signals and behaviors.



\section{Materials and Methods}\label{methods}
\review{In this section we outline the procedures used in our experiments. We first explore the use of LLMs for depression 
classification. In these experiments, we notice that LLMs produce reasoning about the input mobile health data before providing a classification response. We grade this reasoning for numerical accuracy. 
Next, we conduct an interactive evaluation with mental health professionals to gauge the quality of LLM-generated reasoning as it relates to mental health and gain insights into the potential applications for LLM-based tools in therapeutic contexts.}

\subsection{\review{Dataset}}
\review{In this paper, we use the GLOBEM dataset \cite{xu_GLOBEM_2022}, with its extensive collection of passive sensing data from smartphones and wearables and its wide range of well-being metrics. GLOBEM includes weekly Ecological Momentary Assessment (EMA) surveys to capture the recent status of participants' mental health evaluations. 
Within EMA measures, we use Patient Health Questionnaire 4 (PHQ-4)~\cite{kroenke_ultra-brief_2009}} as the ground truth in our depression classification task~\cite{jacobson_passive_2020,huckins_causal_2020,huckins_mental_2020}. Paired with the survey data, GLOBEM contains 24$\times$7 sensing data that measures user behavior, such as steps, GPS locations, phone calls, social activity proxies, and more. The dataset then extracts hundreds of features (time at home, time asleep, etc.) from these raw measurements. 

\subsubsection{\review{Model Input Setup}}
\review{To constrain the input token length, we select a subset of 16 diverse features, including Location, Phone Usage, Bluetooth, Calls, Physical Activity, and Sleep; we list all feature details in Appendix \ref{appendix.data-elements}. We set the time length of the data as 28 days, with the final day of each window coinciding with the weekly EMA assessment.} Therefore, the size of mobile and wearable sensor data for each data sample is 28$\times$16 (28 days $\times$ 16 features). \review{Due to the data protection guidelines for the GLOBEM dataset, we are unable to directly open-source raw excerpts of sensor data or fine-tuned LLMs from which excerpts of training data may extracted\cite{carlini2021extracting}. To aid others in reproducing our results, we provide instructions and code to generate samples prepared from the GLOBEM dataset used in the following experiments in a GitHub repository\footnote{https://github.com/ubicomplab/classification-to-clinical}.}

\subsubsection{\review{Classification Label Setup}}\label{classification-setup}
\review{
We focus on a binary classification task based on PHQ-4 scores, \review{which attempt to quantify the severity of depression and anxiety from a range of 0 (normal) to 12 (severe) based on responses to a questionnaire}. 
As an initial exploration, we avoid the borderline samples with PHQ-4 scores between 1-5, following PHQ-4 criteria~\cite{kroenke_ultra-brief_2009} and supported by additional work finding frequent disagreement between self-reported and clinician-administered scores for borderline cases \cite{ani_depression_2008}. Thus, the classification task aims to distinguish samples with PHQ-4 below 1 or above 5.
To create a balanced test set, 
we randomly sampled 30 data points from each year with an equal distribution of labels to equally represent a range of non-borderline cases.}
In total, the test data set contains 90 class-balanced samples from three years. 

\begin{figure}
    \centering
    \includegraphics[width=\textwidth]{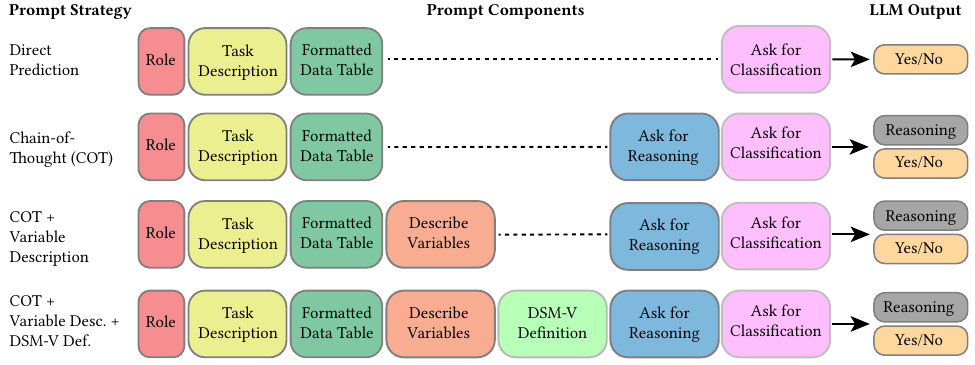}   
    \caption{\review{The overall workflow of our prompting strategies, detailing the specific components included in each type of prompt. For examples of the text included in each block, see Appendix \ref{appendix.prompt-structure}.}}
    \label{fig:prompt}
\end{figure}

\subsection{\review{Classification with LLMs}}
\label{classification-methods}

\subsubsection{\review{Depression Classification}}\label{depression-classification}
\review{We develop prompting strategies to enter the raw data into LLMs along with varying amounts of context and instructions to produce a depression classification result. For each strategy, we include prompt text for the model's role and task description, followed by the concatenated raw sensor data in varying formats, and we proceed to add additional context like variable descriptions and instructions for the task.}\review{For example, a prompt would begin with: \\
\textit{Role: You are a data analyst helping a psychiatrist understand human activity data.\\
Task: You will be shown data gathered from a smartphone and smart watch worn by an individual. Your goal is to analyze this data. You are presented with the following: 1. A table consisting of twenty-eight days of collected activity tracking data <formatted data> ...}}

We show a diagram of each of the utilized prompting formats in Figure \ref{fig:prompt} and a detailed description of the text formatting of each block in Appendix \ref{appendix.prompt-structure}. We focus on three different variants of prompting, all on a zero-shot setting detailed below:

\begin{enumerate}
\item \textit{Direct Prediction (DP):} we directly ask the LLM to perform depression classification with prompts that include only the basic role and task information and the formatted sensor data.
\item \textit{Chain-of-Thought (CoT) prompting:} building upon direct prediction, we induce models to perform step-by-step reasoning with carefully crafted instructions to hypothesize about the subject's overall mental health.
\item \textit{Reasoning with extra information:} based on CoT, we provide extra task-related domain information, such as more detailed explanations of input variables (Exp) and the depression criteria from the Diagnostic and Statistical Manual of Mental Disorders, Fifth Edition (DSM-V)\cite{dsmv-book}.
\end{enumerate}

\review{We test these prompting variants on top of three state-of-the-art LLM models: GPT-3.5, GPT-4 and PaLM 2. We evaluate the DP method with multiple data input formats to determine which is most suitable for the selected LLMs. Specifically, we tried four ways to format these raw data: comma-separated values (CSV), Tabular, Markdown and LaTeX, as shown in Appendix \ref{appendix.data-input-format}. We found that Markdown performed best, so we utilize the Markdown table format in all other experiments.}

\begin{figure}
    \centering
    \includegraphics[width=\textwidth]{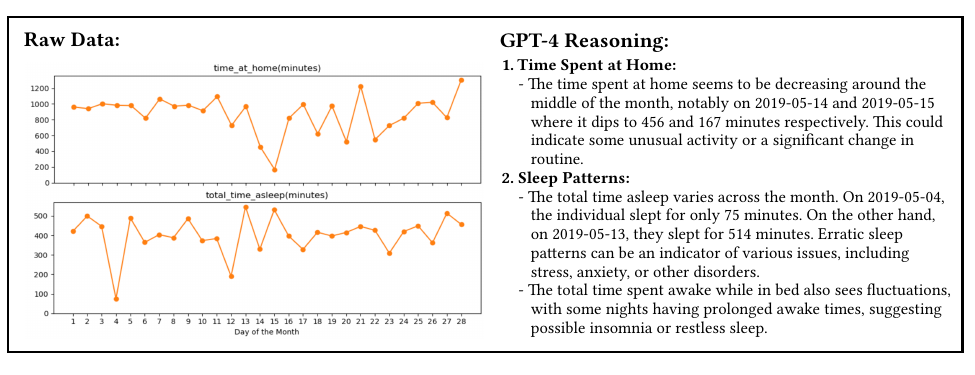}
    \caption{A plotted excerpt of raw mobile health data and the resulting analysis generated by GPT-4.}
    \vskip -0.1in
    \label{fig:analysis-example}
\end{figure}

\review{Using these strategies, we observe that prior to outputting a classification decision, the models first produce intermediate text with reasoning and analysis of the input data. Fig~\ref{fig:analysis-example} shows an example, with a subset of the raw time series input to the LLM plotted for visualization along with a subset of the analysis text produced by GPT-4. These outputs form the basis for our clinician study and envisioned use cases in which these generative LLM outputs can enable new capabilities that go beyond classification to aid clinicians in collaborative therapy. We discuss these extensively in Sections ~\S\ref{sec:reasoning_eval} and \S\ref{sec:clinical_use}.}

\review{To contextualize the classification performance of LLMs, we also introduce} a few baselines, including a classic ML method, Random Forest (RF) \cite{breiman2001random}, as well as the state-of-the-art self-\review{supervised} learning method \textit{Reorder}, proposed by \cite{xu_globem_2023} on this same dataset. Given that these comparison baseline ML models do not support a zero-shot setting and require pre-training, we curated a training dataset for comparative analysis. This dataset comprises the remaining data points that meet the PHQ-4 thresholds,  \review{which excludes the data points in the testing dataset}, totaling 384 samples. These were formatted \review{into two structures: the first is} identical to the test set samples \review{with a structure of 28 days $\times$ 16 features. The second is a set of 16 features computed by taking the average along the time dimension for the Random Forest.}  We conducted evaluations using Random Forest (RF) and Reorder algorithms, alongside the Reorder algorithm \review{with the original implementation in \cite{xu2023globem} with 54 features}  (Reorder-54 Features). These evaluations were performed on the same test set as the LLMs.

\subsubsection{\review{Fine-Tuning}}
Evaluating the CoT reasoning produced by GPT-3.5 reveals that unlike GPT-4, smaller LLMs rarely incorporate analysis that relates specifically to the numerical values of the mobile health data, \review{instead generating generic statements about mental health. Several studies\cite{shridhar2023distilling, yang2023effective, hsieh2023distilling} have shown that the step-by-step CoT reasoning capabilities of larger models can be distilled into smaller models through fine-tuning on the output of larger models.} This raises an important question: can we use instruction fine-tuning to enable smaller LLMs to perform more effectively on such challenging wearable classification tasks? 

To answer this question, we explore using the reasoning responses generated from the GPT-4 \review{(see Appendix \ref{appendix.gpt-4-response})} to fine-tune the GPT-3.5 model. For the fine-tuning experiment, the prompt design was kept consistent with the same methodology outlined above. Utilizing GPT-4, we generate a collection of reasoning responses based on data external to the test set. From this assortment, correctly classified reasoning responses \review{free of numerical errors} were selected to form a candidate training set. We make a balanced instruction training set, comprising 70 sample with high-quality reasoning, evenly distributed between positive and negative examples, \review{and fine-tune GPT-3.5 on this dataset.} 

\subsubsection{\review{Anxiety Classification}}
In addition to depression, we also explore the generalization abilities of LLMs to other mental health classification tasks. Specifically, we use the same CoT prompt design as in the depression classification \review{experiment described in Section \ref{depression-classification}, with the modification of asking for anxiety classification based on the PHQ-4 anxiety sub-score instead of the depression classification.}

\subsection{\review{Evaluating Numerical Accuracy}}

We begin by investigating the \review{accuracy of the LLMs when referencing the input data.} Can the models identify real and specific trends in the data? Do they hallucinate numbers, or are the outputs truly reflective of the input data? To evaluate this, we utilize human graders to \review{objectively evaluate the numbers and stated trends referenced in }LLM responses against the input timeseries data according to a fixed rubric. We score a total of 480 responses, evenly split across four different models: PaLM 2, GPT-3.5, fine-tuned GPT-3.5, and GPT-4.


\subsubsection{Producing Reasoning Samples} \review{\label{producing-reasoning-samples}}Instead of asking to hypothesize about the health of the patient, we tune this prompt to produce analysis on trends in the data. \review{We show the specific prompt used in Appendix \ref{appendix.reasoning-prompt}.} For input data, we select random samples from the test set with a PHQ-4 score of greater than 5, indicating likely moderate to severe depression. We select these samples since  they are most likely to have a trends or anomalies on which we can evaluate the model. These excerpts consist of the same 16 features used in Section \review{\ref{classification-methods}} formatted in markdown format. We produce 8 samples from each of the four models per set of input data. 

\subsubsection{\review{Participants}}
\review{To grade reasoning excerpts, we recruited 15 individuals over the age of 18 through fliers placed around a university campus and word of mouth. Participants were not required to have any domain knowledge of mental health to complete the assigned task. To aid in recruitment, compensation of a gift card (\$20 USD) was offered to all participants included in the study. The study protocol was submitted to the IRB at the host institution for the study and deemed exempt from a full review. We follow all IRB procedures to avoid potential conflict of interest.}

\subsubsection{\review{Prompt Grading Procedure}} Each participant was sent an online form that contained a set of 32 randomly ordered responses (8 from each LLM). To reduce grader burden, all 32 responses were generated from the same input data. Graders were also provided the raw data input to the model in tabular form as well as timeseries plots of each data  feature. For each reasoning excerpt, graders were asked to answer the following four questions:
\begin{enumerate}
    \item Does this response include numbers?
    \item Are these numbers consistent with the provided data?
    \item Does this response identify specific trends?
    \item Are these trends consistent with the provided data?
\end{enumerate}

Graders were given specific instructions on how to evaluate each question as well as a series of \review{graded example responses}; we show the rubric provided to graders in Appendix \ref{appendix.rubric}. We provide explicit instructions to evaluate \review{solely the numbers and trends against the provided data table and plots, disregarding any conclusions the responses may make about how these numbers or trends might relate to mental health or other factors.} An excerpt of plotted data and LLM-generated reasoning graded by participants is shown in Fig. \ref{fig:analysis-example}.

\subsection{\review{Clinician Evaluation}}
\label{sec:reasoning_eval}
As shown in Fig~\ref{fig:analysis-example}, the LLMs also generate analyses about the timeseries data. To evaluate the quality of this reasoning and understand how these clinical insights could be used in practice, we conducted a user study with clinician experts.

\subsubsection{Participants}
Eight mental health professionals completed this study: six with PhDs in Clinical Psychology and two with master's degrees. Participants described their approaches as Cognitive-Behavioral Therapy (2), Acceptance and Commitment Therapy (1), Dialectical Behavioral Therapy (1), Psychodynamic Therapy (1), Relational/Interpersonal Therapy (2), and Family Systems Therapy (1). They reported working in a variety of settings, including academic medicine, group private practice, individual private practice, and community mental health, across four states in the U.S. \review{Participants were recruited through postings on group practice mailing lists, social media groups for practicing therapists, and word of mouth. To aid in recruitment, gift cards (\$50 USD) were offered to all who  participated in the study.}

\subsubsection{\review{Ethics}}
\review{This study protocol was submitted to the IRB at the host institution and was deemed exempt from formal IRB review. Participants were sent an information sheet outlining the study process and data management procedures prior to interviews. At the start of the interview but before initiating recording, researchers described the interview process, addressed questions, and obtained spoken consent.} 

\begin{figure}[h]
    \centering
    \includegraphics[width=0.8\textwidth]{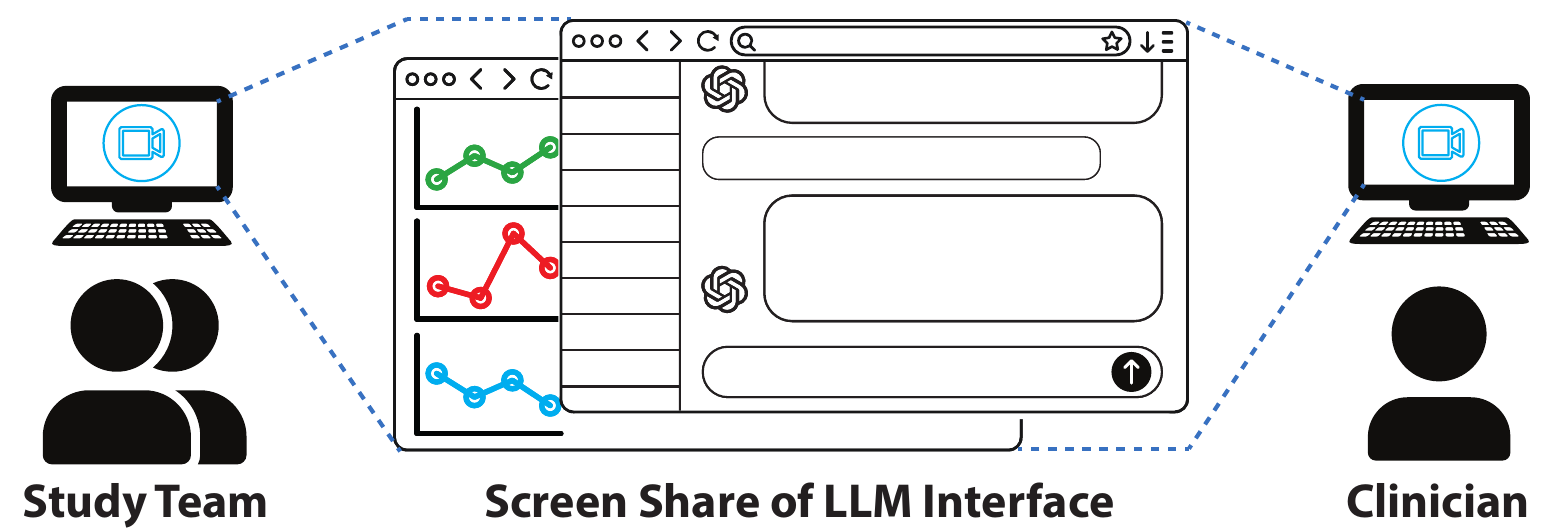}
    \vskip -0.10 in
    \caption{Setup for interactive clinician evaluation.  
    Clinicians interacted with live GPT-4 sessions via screen sharing~\review{over Zoom.}}
    \label{fig:clinician-setup}
    \vskip -0.2in
\end{figure}

\subsubsection{\review{Data Format}}\review{The data analyzed in each interview session consisted of a random 28-day sample from a participant in the GLOBEM dataset~\cite{xu_GLOBEM_2022}. Only individuals with a PHQ-4 $>$ 5 were selected (as in Section \ref{classification-methods}) to obtain a sample representative of individuals who might be likely to seek therapy. 
Time-series plots that included each variable were generated.  
The generative reasoning evaluation drew on the same 16 features and format used for the time series (listed in Appendix A). The chat thread was started with the following prompt, similar to the prompt used in the previous section, but altered slightly for compatibility with the web-based chat interface of GPT-4: \textit{Below is some data gathered from a fitness tracking smartwatch and a smartphone. Although it does not contain explicit information on mood, trends in physiological signals have been shown to correlate with mental health symptoms. Examine this data and point out any specific trends or data points that could spark fruitful conversation with a mental health professional. <formatted-data>}. }

\subsubsection{Interview \review{and Interactive Exercise}}
Interviews were one hour in length and conducted over Zoom to enable recording and transcript generation. \review{Participants were not shown the results of other experiments conducted in this paper.} Interviews were conducted by two researchers, a clinical psychologist and a computer science graduate student.

Interviews began with a discussion of the clinician’s practice, particularly as it includes patient self-monitoring. We then asked participants to imagine a scenario in which they had received a month of self-tracking data in advance of their first meeting with a patient. To build on this scenario, we shared examples of self-tracking data from the same publicly available data set used above~\cite{xu_globem_2023} in two forms. First, we presented a set of 16 time series plots (in a shared Google doc), and clinicians were asked to provide feedback on anything they noticed in the data that they might use in a therapeutic context. 

Next, we began a live interactive session with GPT-4 (using ChatGPT) through Zoom screen sharing, as shown in Figure~\ref{fig:clinician-setup}. The mobile health data was then input into GPT-4 to produce a text response in real-time. We asked participants to read the response and talk about any reactions they had, including ways the data and GPT-4’s observations about the data might shape their thinking about the patient. Participants were asked to type or say aloud follow-up queries that they had for GPT-4 (which were entered by the interviewer running the GPT-4 session). We also discussed their reaction to the responses that GPT-4 gave to these queries. We then asked participants to make up a hypothetical example in which they used this tool with a therapy patient, prompting them for what types of data or inputs should be available, what kinds of analyses they would like to see, and how they would envision using this tool with the patient. Finally, we asked general questions about treatment and how this tool might affect relationships with patients. 

\subsubsection{\review{Post-Interview Survey}}
After the interview, participants completed a short online survey 
where they indicated their agreement or disagreement with  several statements \review{about GPT-4  on a seven-point scale.}


\review{\section{Results}}
In this section, we provide results of the depression classification, fine-tuning, and anxiety classification experiments as well as the numerical accuracy and clinician evaluations, all described in Section \ref{methods}.

\subsection{\review{Depression Classification}}
\begin{figure}[h]
    \centering
    \includegraphics[]{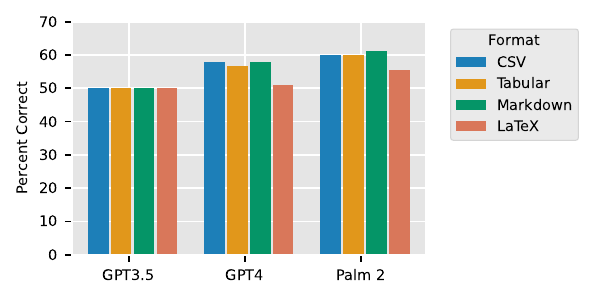}
    \vskip -0.25in
    \caption{Depression classification performance across a range of possible data input formats.}
    \label{fig:format_results}
\end{figure}

\review{To evaluate multiple data input formats, we use the classification results of depression through Direct Prediction (DP) as a measure of these formats. From Figure \ref{fig:format_results}, we observe that CSV, Tabular and Markdown formats exhibit comparable performance levels,  and these results are consistent in both GPT-4 and PaLM 2. In contrast, the LaTeX format demonstrates a performance gap compared to the other three formats. These results align logically with expectations, considering the predominant sources of training data for LLMs. The vast majority of this data is sourced from the Internet, where formats like CSV and Markdown are far more prevalent than LaTeX. Given this disparity in data availability, it stands to reason that LLMs would exhibit higher accuracy in processing and interpreting CSV and Markdown inputs compared to LaTeX. Since Markdown shows the best overall performance, we chose Markdown as the data format for our subsequent experiments.}

\begin{figure}[t]
    \centering
    \includegraphics[width=\textwidth]{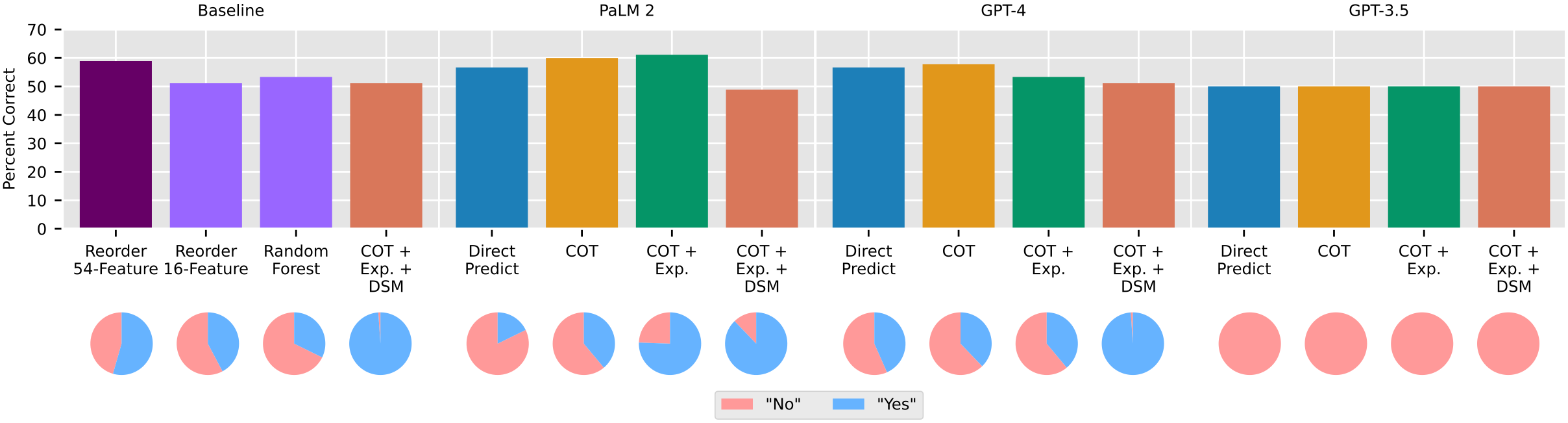}
    \caption{\review{Comparison of classification results of GPT-3.5, GPT-4, and PaLM 2 across four different prompting strategies, along with results from Reorder and Random Forest models trained on the same dataset. The performance of the Reorder model trained on more features in ~\protect\citet{xu_globem_2023} is included for comparison.} Observe how for LLMs, the percent of positive ("Yes") and negative ("No") classifications varies significantly based on the prompting strategy used. }
    \label{fig:depression-classification}
\end{figure}

We tested all prompting strategies with three state-of-the-art LLM models, GPT-3.5, GPT-4 and PaLM 2 in Markdown format. Figure \ref{fig:depression-classification} reveals that while GPT-3.5 attains an accuracy rate of 50\%, it does not effectively address the question posed. Instead, it consistently defaults to a response of 'No', which results in an inflated 50\% accuracy due to our balanced dataset. Figure \ref{fig:depression-classification} shows that CoT improves the accuracy of both GPT-4 and PaLM 2 compared to DP, \review{which aligns with the results} found in many related studies \cite{wei_chain--thought_2023, wang2022self}. Using the CoT + Exp. strategy, PaLM 2 achieved the highest accuracy of 61.11\%.
However, adding information, even if accurate and pertinent to the topic, does not always increase performance. Both PaLM 2 and GPT-4 perform their worst, at 48.89\% and 51.11\%, respectively, when provided with the DSM-V description of depression as part of the prompt. In these cases, we see that this results in a significant increase in the percentage of samples classified as depressed, with GPT-4 classifying as high as 98.89\% of samples as positive. 

The performance metrics reveal that the \review{accuracies of Random Forest}, Reorder, and Reorder-54 Features are 53.33\%, 51.11\%, and 58.89\%,  respectively. Notably, among the two baseline models utilizing the same set of 16 features \review{provided to the LLMs, neither} could surpass the zero-shot Chain of Thought (CoT) results achieved using the GPT-4 and PaLM 2 models.
\subsection{\review{Fine-Tuning}}

\begin{figure}
    \centering
    \includegraphics[]{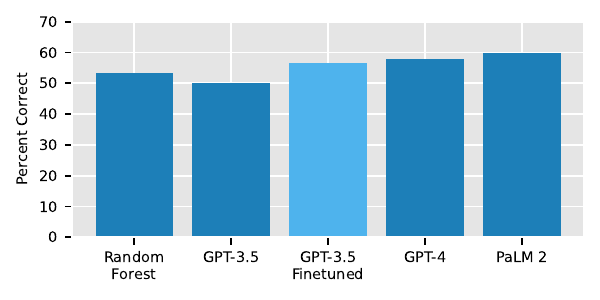}
    \vskip -0.25 in
    \caption{Comparison of \review{depression classification performance for} GPT-3.5 fine-tuned for chain-of-thought (CoT) reasoning \review{depression classification} against CoT performance of \review{GPT-3.5, GPT-4, and PaLM 2. The performance of the Random Forest model is included as a baseline.}}
    \label{fig:gpt-35-ft}
    \vskip -0.2 in
\end{figure}
After fine-tuning GPT-3.5 with 2 epochs using this balanced instruction-tuning set, we see an improvement in its performance.  Though it initially fails to properly perform classification, our fine-tuned version of GPT-3.5 achieves an accuracy of 56.67\% on the test set, \review{which exceeds the baseline Random Forest model performance of 53.33\% and is closer to the GPT-4 performance of 57.78\% on the same dataset and prompt structure.} 
\subsection{\review{Anxiety Classification}}
\begin{figure}
    \centering
    \includegraphics[]{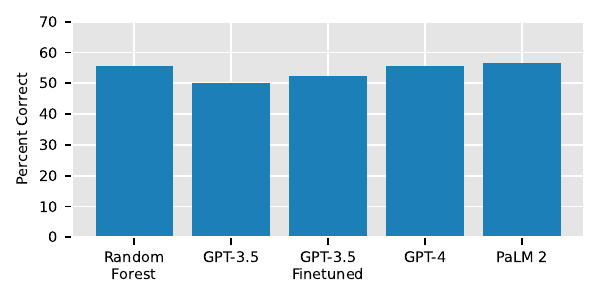}
    \vskip -0.2in
    \caption{\review{Comparison of anxiety classification performance for GPT-3.5 fine-tuned for chain-of-thought (CoT) reasoning depression classification against CoT performance of GPT-3.5, GPT-4, and PaLM 2. The performance of the Random Forest model is included as a baseline.}}
    \label{fig:anxiety}
\end{figure}
As in the depression classification tasks, we observe that GPT-3.5 still always defaults to a "No" response. GPT-4 can achieve an accuracy of 55.56\%, while PaLM 2 achives a slightly higher accuracy of 56.67\%. Notably, this result mirrors the trend observed in the depression classification results. For our fine-tuned GPT-3.5 model, although the model is fine-tuned for depression classification, it still shows some improvement in the anxiety detection task compared to the original GPT-3.5. This indicates the potential of fine-tuning to increase performance on classification tasks. 

\begin{figure}
    \centering
    \includegraphics[]{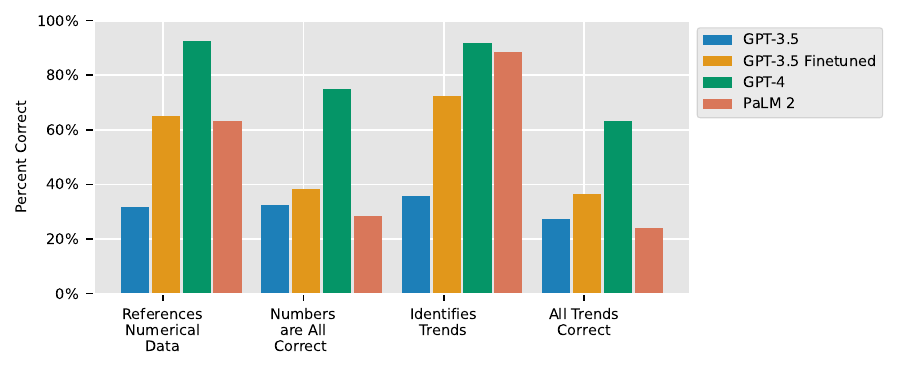}
    \vskip -0.2in
    \caption{Evaluation of the reasoning excerpts generated by GPT-3.5, Fine-tuned GPT-3.5, GPT-4, and PaLM 2.}
    \label{fig:grading}
\end{figure}

\review{\subsection{Evaluating Numerical Accuracy}}
Figure~\ref{fig:grading} shows the results of our study. Interestingly, we find that while PaLM 2 performs slightly higher on the classification tasks above, GPT-4 performs significantly better across our study rubrics. For example, while PaLM 2 also identifies trends at a high rate, GPT-4 is more likely to identify all trends correctly and include references to the numerical data, achieving scores exceeding 75\% accuracy. We further note that the graders evaluated data as simple yes/no questions, meaning that all numbers and trends had to be correct. Although we observe some errors in the model outputs, even these responses often contain correct trends as well that \review{may have utility} in a collaborative human-AI approach. For example, if the LLM can observe an outlier or anomaly in the data, this is often easily visible to the user as well for confirmation. 

\review{\subsection{Clinician Evaluation}}

\begin{figure}
    \centering
    \includegraphics[width=\textwidth]{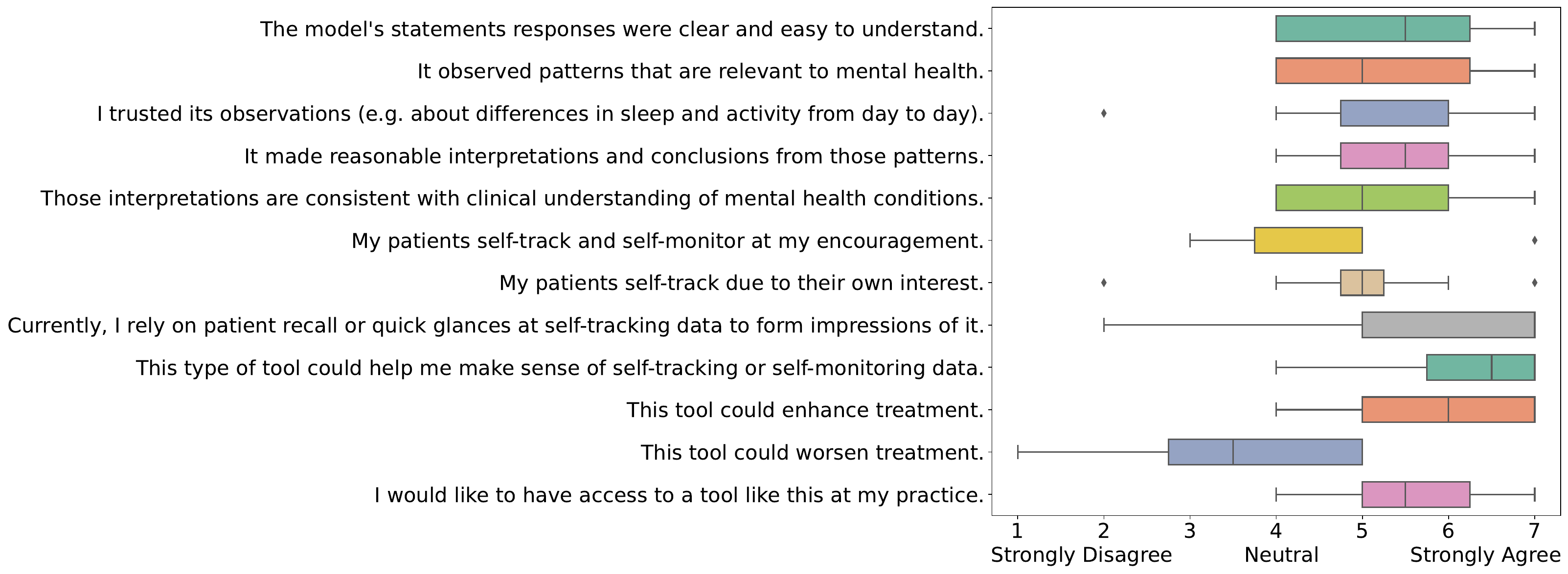}
    \vskip -0.1in
    \caption{Post-study survey responses from clinicians, indicating strong disagreement (1) to strong agreement (7) with statements about their experience interacting with GPT-4. We observe positive feedback and enthusiasm across most questions.}
    \label{fig:clinician-survey}
\end{figure}

\review{We discuss findings from clinical interviews in two categories. First, we describe envisioned usages, that is how clinicians thought about incorporating the tool into their practice. Second, we describe concerns that came up as clinicians envisioned using the tool. We preface these findings by noting a general interest among clinicians in having access to LLM-based tools in their practice (see the post-study survey responses in Fig. \ref{fig:clinician-survey}}.
\\
\subsection{Envisioned Clinical Uses}\label{sec:clinical_use}
\review{Three major uses were envisioned by clinicians: collaborative investigation, identifying questions to explore, and documentation. }

\subsubsection{Collaborative In-Session Investigation}
Clinicians generally saw the most value in this tool to aid collaboration with patients by using it as an \textit{interactive data explorer during a therapy session}. For example, one clinician imagined querying the model to identify triggers for panic attacks and other anxiety symptoms. Another clinician outlined the high-level steps she would use if incorporating the model: \textit{“First, set the goal... What's bringing them in? Then, agree on some metrics that are relevant.”} (P4). 
Others imagined using it to \textit{find evidence of change early in treatment }(e.g., indications of more energy or better sleep) as a patient struggled with whether to continue a particular medication or therapy. One therapist gave an example of how she might, through discussion with a patient, tie a concern such as relationship anxiety to the average duration of phone use periods and then use that metric to assess whether therapy was helping with the patient’s anxiety. Several imagined using it not only for retrospective analysis but also \textit{to forecast improvement}. Another envisioned use was \textit{asking the model questions during sessions to boost creativity}, e.g., when challenging a patient’s worries or negative thoughts. 

Across these and other forms of collaborative use, \textit{some clinicians wanted their patients to be able to query the model}. In addition to identifying patterns, joint use was envisioned as a way to build a more general feeling of collaboration in the therapy, something that is important for the therapeutic alliance and positive outcomes~\cite{martin2000relation,cameron2018relationship}.

\subsubsection{Identifying Issues to Explore with the Patient}
Clinicians appreciated the model's ability to list concerns with references to specific dates (e.g., days with little sleep or little movement). One clinician shared that it would be useful for GPT-4 to prioritize the questions. Clinicians suggested that they would share their observations with patients, e.g., about particular days with anomalies such as decreased sleep, as a way of opening up discussions and jogging the patient’s memory. 


When discussing potential concerns, clinicians wanted the model to raise questions and supply clear data. As one clinician said, the model "Could cause less harm if it provided questions for a therapist to ask instead of conclusions for a therapist to rely on" (P8). 
They did not want the model to apply diagnostic labels to individuals or their behaviors. 

\subsubsection{Generating Documentation}
Clinicians varied in their thoughts on whether models would meaningfully aid in documentation. One participant appreciated the neutral boilerplate language the model used and easily imagined it as the basis for an intake summary. \review{Another had already started using AI software to aid in documentation of sessions.} Others bristled at this idea, pointing out that the LLM might miss major insights and obviate the analysis that comes from writing notes.


\subsubsection{\review{Concerns}}
\review{Clinicians raised additional concerns related primarily to privacy and quality of care.}

\textbf{Privacy. }\review{As clinicians considered inputs, such as patient mood logs, that would increase the model's relevance to mental health, they grappled with privacy and other ethical concerns. One clinician worried that it could be hard to obtain meaningful informed consent. Even if a model were secure and compliant with relevant patient privacy regulations, it could be difficult for therapists to explain how the system worked and what protections it afforded.}

\textbf{Quality of Care. }Another concern pertained to relying on the model as a shortcut. Clinicians imagined the problems that could arise from overconfidence in the model’s analysis. One clinician described the possibility of missing the insights that would come from pouring over data herself: \textit{“If I was...doing this very quickly, like, before I see the client, I could totally see myself or anyone...just relying on on GPT-4. And just think … 'This is the answer. This is the knowledge,’ ... as mental shortcuts as opposed to pouring over the data yourself. … And I would want to ask, ‘Am I missing anything?’} (P6). Another clinician 
worried that not taking the time to write one’s own notes and process each session could degrade the their memory of the session and ultimately shortchange the therapy. \review{And 
contrasted her efforts to examine contextual factors, such as family conflict, with a model's narrower focus on symptoms. She noted that if relying solely on a model's observations, a clinician could overlook important factors in assessing and treating mental health struggles.}

\section{discussion}
\review{
Our findings reveal a paradigm shift based on the value LLMs can bring to mental health care, switching from a focus of using computational tools for diagnostic classification to using them to generate clinical reasoning. Most prior ML research in mental health has focused on classification, or predicting the diagnostic category determined by a clinical interview or score on a self-report inventory such as the PHQ. While these coarse categorizations have a role in mental health care, they do not capture the specifics of a particular patient's struggles and are insufficient to meaningfully guide treatment. Therapeutic change relies on the patient becoming aware of their patterns and options for changing them: it is an active collaboration rather than a procedure performed on a patient based on a particular diagnosis. Further, therapy is hyper-personalized, grounded in each patient's specific concerns, goals, contexts, and dynamics.} 

\review{Our findings indicate that LLMs have the potential to illuminate an individual's patterns by analyzing disparate sensor data that is otherwise unwieldy for clinicians to interpret. Unlike a diagnostic wizard used before treatment, LLMs can \textit{bring value in-session} and \textit{throughout treatment}. Instead of being used solely by the clinician or another professional in a care system, LLMs can be \textit{used collaboratively by the patient and clinician}.  In this vision, the LLM does not replace the clinician or replace dialogue between the clinician and patient. Instead, the LLM \textit{facilitates their dialogue}, the alliance that supports change, and their investigation of the patient's patterns. }

Below, we first discuss the challenges of using LLMs for classification followed by further steps needed to develop end-to-end tools for the proposed clinician-patient-AI collaborative therapy. 

\subsection{\review{Challenges of Depression Classification}}
\review{This study illuminated new approaches to and challenges of screening for depression based on mobile health data. Our best-performing method for classifying depression was incorporating chain-of-thought (COT) prompting and variable explanations with the PaLM 2 model, achieving an accuracy of 61.11\% on our class-balanced dataset. Although this represents a modest improvement over classical ML methods such as Random Forest and the state-of-the-art Reorder \cite{xu_globem_2023} method on this dataset, \textit{this level of accuracy is still far too low to be useful as a clinical screening tool.} It is important to note that personalized depression forecasting approaches where model training data includes historical data from test set individuals report accuracy as high as 90.1\% \cite{lee_prediction_2023}, but a reliance on extensive prior labeled data for each given individual significantly limits real-world utility. }

\review{
In our experiments, we utilized a class-balanced dataset, as described in Section \ref{classification-setup}. This provides an indication of performance across an even distribution of individuals with potential PHQ-4 scores. With an unbalanced dataset constructed from all complete samples in the Globem dataset, the distribution would include 34.8\% positive samples as opposed to 50\%. With this distribution, we would expect to see a shift in the optimal prompting strategies, improving performance for direct prediction and COT prompting methods more likely to classify samples as not depressed and further decreasing performance of prompting strategies, including DSM-V depression criteria.}

\review{We also attempted experiments to predict severity through a variety of methods, such as asking the LLMs to select a PHQ-4 score between 0 and 12 or options ranging from "none" to "severe" with varying degrees of additional context. In these instances, though, we notice that all three LLMs always select the middle value of the provided range. This result aligns with the findings of other works, which show that GPT-3.5 and GPT-4 both require few-shot prompting with multiple labelled input examples to perform these types of regression tasks \cite{kim_health-llm_2024}. We forego few-shot prompting experiments for this work since they would require either the individual to track PHQ-4 scores as ground-truth, which is the variable we are trying to predict with the LLM in the first place, or use labelled data excerpts from other participants, which raises the challenging question of how to source representative samples from existing datasets to use for a new individual. }

\subsection{\review{Managing Inaccuracy in a Clinical Context  }}
\review{Clinicians in our study expressed a strong preference for LLMs to generate observations and insights about data relating to a patient rather than apply diagnostic labels. Such usage of  LLM-based tools for generative reasoning avoids the challenges of inaccurate classification, but inaccuracies still need to be managed. 
In particular, well-documented numerical inaccuracies 
may impact the quality of the generated reasoning. In our study, reasoning responses generated by GPT-4, based on a 448-element table (16 features x 28 days), contain at least one numerical error 25\% of the time. The necessary accuracy and tolerance for inaccuracy may vary with applications of the LLM. For example, an 
inaccuracy in a question that is generated for the therapist to ask a patient may pose less risk 
than an error in a definitive statement that informs a clinician's impressions of a patient's mental state. These risks of inaccuracy may be mitigated by uses that prioritize the therapeutic relationship and the patient's perspective. The clinicians in this study anticipated that they would use LLM-based tools in collaboration with patients, 
drawing on the tool's observations as a starting point for discussion. They emphasized that their treatment decisions and diagnoses would be based on their direct patient interactions. }

\subsection{\review{Supporting Clinician-Patient-AI Collaboration}}\label{supporting-collaboration}
\review{Supporting the kind of collaboration we propose between clinicians, patients, and AI will require new approaches to LLM hosting and data protection. This is due in part to the sensitivity of the data required. }To inform mental health treatment and illuminate the factors associated with a particular individual's struggles, models need to have data that are directly relevant to mental health and, ideally, personalized to individual patients. The clinicians we interviewed expected the model to have, at a minimum, daily mood and symptom tracking and ideally more in-depth data related to mood, social interactions, and behavioral routines. In addition, 
a clinician may want to input \review{materials} such as session notes or transcripts, manuals, or other documents describing relevant mental health issues and treatments.  Much of this \review{potentially personally identifiable} data is sensitive and not appropriate as an input for \review{existing LLM services such as ChatGPT, which may use this data to improve their models or for other commercial purposes.} \textit{Privately hosted models or robust user data protections are required for this purpose.}

The solution to the privacy challenges of using LLMs in therapy is not as straightforward as it is for compliant medical record systems. Such record systems primarily serve providers and other employees at medical institutions, limiting patients to read-only access to elements such as test results. In this study, we heard from clinicians that both patients and clinicians should be able to interact with the model. This joint interaction, clinicians anticipated, could build patients' curiosity about factors associated with their mental health and foster the collaborative alliance between patients and therapists that is associated with positive outcome~\cite{martin2000relation, cameron2018relationship}. 

Such models therefore require a very different approach than that used for medical records; in this case, \textit{the patient should own their data, but both parties (the patient and therapist) should be able to generate data and actively interact with it. }\review{Meeting the computational requirements of hosting such services may be challenging for providers in smaller practices, while relying on third-party services raises questions about data ownership. Addressing these concrete privacy challenges will open the door for broader exploration. \textit{AI systems intended not just to produce a single diagnosis but rather to engage users in collaborative investigation over time will require new modes of interaction.} This will pose research and design challenges in designing systems for use in mental health care and other contexts.}

\subsection{\review{Limitations and Future Work}}

\subsubsection{\review{Dataset Generalization and Balancing}} \review{In our experiments, we utilize a class-balanced dataset confined to a subset of mobile and behavioral health data collected from undergraduate students at a single university over a three-year period. Future evaluations should seek to include both a broader range of data elements as well as data from more diverse population samples to better understand how well these approaches apply to any potential individual seeking mental health treatment. }

\subsubsection{\review{Model and Prompting Bias}} \review{In addition to issues of population representation in the dataset, a growing body of work suggests that state-of-the-art LLMs are prone to producing output containing both implicit and explicit biases \cite{dong_probing_2023, shaikh_second_2023}. This is further complicated by the sensitivity of classification performance to the specific information included in the prompt. For example, providing the DSM-V depression criteria results in GPT-4, which one may assume may improve depression classification, classifying 98.89\% of data excerpts as depressed despite a class-balanced dataset. It is almost certain that prompt and model-specific biases exist in the generated reasoning about the provided mobile health data as well, but this is a significantly more complex problem to characterize. Benchmarking techniques are actively being explored in this area \cite{yeh-etal-2023-evaluating, parrish_bbq:_2022}, and further work to understand how these biases arise and can be mitigated are essential before deployment of LLMs mental health contexts. }

\subsubsection{\review{Numerical Errors}} \review{Clinicians envisioned inputting both a wide range of additional features as well as a longer time period of data into LLMs for analysis. However, the maximum context window of these models is limited, and prior work has shown that the ability of LLMs to accurately recall detailed information from the entirety of the context window decreases as prompt length increases\cite{liu_lost_2023}. Pre-computing or summarizing portions of data could help address this technical challenge, but requiring this step could significantly reduce the ability of a tool to generalize across potential input data sources. }

\review{Some errors observed by clinicians involved GPT-4 generating erroneous responses when asked follow-up questions that could not be answered based on the provided mobile health data. This challenge of \textit{grounding}, or constraining LLM output to refer only to a relevant set of input text or data, is not unique to this specific application. Current approaches to improve performance in these areas include model fine-tuning\cite{azerbayev_llemma:_2023}; augmenting a base general-purpose LLM with a smaller model optimized for a specific task\cite{bansal_llm_2024}; and utilizing LLMs capable of querying external databases and tools to request data and perform arithmetic operations\cite{gao_pal:_2023}. We view this third option as especially promising since enabling a model to access data from a user-controlled database may simultaneously help address the privacy challenges identified in Section \ref{supporting-collaboration}.}

\subsubsection{\review{Evaluation in Clinical Settings}} \review{While we find clinicians were frequently able to identify errors generated by GPT-4 when they arose during our interviews, in an actual therapy session the clinician may not be able to balance the demands of critically evaluating model output while engaging with the patient. Additionally, our current work solely evaluates LLM use in therapy from the perspective of clinicians. Further evaluation in a more realistic therapeutic environment is needed to better understand how use of LLMs as part of a mental health treatment program could impact patients and their  outcomes. 
}
\section{conclusion}
\review{This paper examines LLMs in the context of mental health care, specifically psychotherapy. 
While we begin our investigation by developing methods to use LLMs to output binary classifications for conditions like depression, we find instead that their greatest potential value to clinicians lies not in diagnostic classification, but rather in rigorous analysis of diverse self-tracking data to generate natural language summaries synthesizing multiple data streams and identifying potential concerns. Clinicians envisioned using those insights in a variety of ways, principally for fostering collaborative investigation with patients. This collaboration was seen as potentially valuable for strengthening the therapeutic alliance and guiding treatment. 
We describe a human-AI collaborative model and its requirements for secure management of personal data. These findings highlight directions for impactful future research on human-AI collaborative tools in mental health care and other contexts. }
\begin{acks}
This research was partially supported by an Amazon Research Award, a Google Research Scholar Award, The Washington Research Foundation, and the Pastry-Powered T(o)uring Machine Endowed Fellowship.
\end{acks}



\appendix

\section{Data Elements}\label{appendix.data-elements}
\begin{table}[H]
\centering
\begin{tabular}{|l|l|}
\hline
\textbf{GLOBEM Data Feature} & \textbf{Description} \\ \hline
date & date \\ \hline
f\_loc:phone\_locations\_doryab\_totaldistance:allday & total distance traveled (meters) \\ \hline
f\_loc:phone\_locations\_doryab\_timeathome:allday & time spent at home (minutes) \\ \hline
f\_loc:phone\_locations\_doryab\_locationentropy:allday & location entropy \\ \hline
f\_screen:phone\_screen\_rapids\_sumdurationunlock:allday & phone screen time (minutes) \\ \hline
f\_screen:phone\_screen\_rapids\_avgdurationunlock:allday & average phone unlock duration (minutes) \\ \hline
f\_call:phone\_calls\_rapids\_incoming\_sumduration:allday & phone call incoming duration (minutes) \\ \hline
f\_call:phone\_calls\_rapids\_outgoing\_sumduration:allday & phone call outgoing duration (minutes) \\ \hline
f\_blue:phone\_bluetooth\_doryab\_uniquedevicesothers:allday & unique Bluetooth devices discovered nearby \\ \hline
f\_steps:fitbit\_steps\_intraday\_rapids\_sumsteps:allday & step count \\ \hline
f\_steps:fitbit\_steps\_intraday\_rapids\_countepisodesedentarybout:allday & number of sedentary episodes \\ \hline
f\_steps:fitbit\_steps\_intraday\_rapids\_sumdurationsedentarybout:allday & total time spent sedentary (minutes) \\ \hline
f\_steps:fitbit\_steps\_intraday\_rapids\_countepisodeactivebout:allday & number of activity episodes \\ \hline
f\_steps:fitbit\_steps\_intraday\_rapids\_sumdurationactivebout:allday & total time spent active (minutes) \\ \hline
f\_slp:fitbit\_sleep\_intraday\_rapids\_sumdurationasleepunifiedmain:allday & total time asleep (minutes) \\ \hline
f\_slp:fitbit\_sleep\_intraday\_rapids\_sumdurationawakeunifiedmain:allday & total time spent awake while in bed (minutes) \\ \hline
\end{tabular}
\caption{Data Fields and Descriptions}
\label{table:data-fields}
\end{table}

\section{Prompt Structure}\label{appendix.prompt-structure}
Each following subsection outlines the specific text used for each component of the prompt. To re-create our prompts for a given experiment, concatenate the relevant sections as shown in Fig. \ref{fig:prompt}. 
\subsection{Role}
\begin{verbatim}
Role:  
You are a data analyst helping a psychiatrist understand human activity data. 
\end{verbatim}
\subsection{Task Description}
The numerically listed items in this block should be added or removed according to the blocks included in the rest of the prompt. 
\begin{verbatim}
Task: 
You will be shown data gathered from a smartphone and smart watch worn by an individual. Your 
    goal is to analyze this data.
You are presented with the following: 
1. A table consisting of twenty-eight days of collected activity tracking data [Collected Data] 
2. Explanation of the different types of data. [Description of Variables]
3. DSM-5 Criteria for depression. [Depression Criteria]
4. Instructions on how to analyze the data [Instructions]
\end{verbatim}
\subsection{Formatted Data Table}
Begin with a header \textit{Collected Data:} followed by the data. Each of the data formats used in this paper can be found in Appendix \ref{appendix.data-input-format}.
\subsection{Describe Variables}
These descriptions may vary based on the specific dataset used. 
\begin{verbatim}
Description of variables
For the duration of the data collection period, data was logged from the individual’s 
    smartphone and a fitness tracking smartwatch. An explanation of the different 
    types of data logged is included below:
Date: the data in this row was recorded on that date.
total_Distance_traveled(meters): This represents the total distance traveled as measured
    by the GPS in the participant’s smartphone. This includes walking, driving, and any 
    other modes of transportation.
time_at_home(minutes): This is the total time that the participant was at home as 
    determined by their smartphone GPS.
…
\end{verbatim}

\subsection{DSM-V Definition}
\begin{verbatim}
Depression Criteria
Here is the DSM-5 Criteria for depression. The DSM-5 outlines the following criterion to 
    make a diagnosis of depression. The individual must be experiencing five or more symptoms 
    during the same 2-week period and at least one of the symptoms should be either 
    (1) depressed mood or (2) loss of interest or pleasure…
\end{verbatim}
\subsection{Ask for Reasoning and/or Classification}
Specify the type(s) of analysis required. The example below includes both reasoning and classification instructions.
\begin{verbatim}
Instructions 
1. Based on the collected data, hypothesize about the overall mental health of the individual 
    with detailed explanations.
2. Make your best guess if this person is currently experiencing depression. Please answer 
    only "Yes" or "No" to this question.
\end{verbatim}

\section{Example GPT-4 Response}\label{appendix.gpt-4-response}
Below is a response generated by GPT-4 from a prompt consisting of the Chain of Thought, Variable Description, and DSM-V Definition elements:
\\
\begin{center}\rule{0.5\linewidth}{0.5pt}\end{center}

To assess the mental health of the individual, we'll analyze the
provided data and relate it to the DSM-5 criteria for depression.
Hypothesis About Overall Mental Health:

\begin{itemize}
\item
  \textbf{Physical Activity}: The individual's step count fluctuates
  significantly, with days of very high physical activity (e.g., 55,755
  steps on 2019-05-11) and others with much lower activity (e.g., 2,050
  steps on 2019-05-20). The inconsistency in physical activity could
  indicate variability in energy levels or mood.
\item
  \textbf{Time at Home}: There are significant variances in the amount
  of time spent at home. Some days show extended periods at home (e.g.,
  1,240 minutes on 2019-05-12), while others show much less (e.g., 5
  minutes on 2019-05-19). Spending extended periods at home might
  suggest a lack of engagement in external activities, possibly
  indicating social withdrawal or lack of interest.
\item
  \textbf{Phone and Screen Usage}: The phone screen time and average
  phone use unlock duration fluctuate, with some days showing high usage
  (e.g., 819 minutes on 2019-05-21) and others considerably lower (e.g.,
  138 minutes on 2019-05-19). This could indicate variations in the need
  for distraction or communication.
\item
  \textbf{Sedentary and Active Episodes}: There's inconsistency in the
  number of sedentary and active episodes. Some days have high sedentary
  time and low activity (e.g., 1,402 sedentary minutes vs.~38 active
  minutes on 2019-05-20), which might suggest fatigue or lack of energy.
\item
  \textbf{Sleep Patterns}: The total time asleep and awake while in bed
  varies. There are nights with less sleep (e.g., 249 minutes on
  2019-05-10) and others with more (e.g., 679 minutes on 2019-05-12).
  Irregular sleep patterns could be indicative of disturbed sleep, which
  is a symptom of depression.
\item
  \textbf{Location Entropy}: On most days, location entropy is 0,
  suggesting a lack of diversity in location visits. This could imply a
  routine life with limited exploration or lack of interest in diverse
  activities.
\item
  \textbf{Phone Call Duration}: There is limited data on phone call
  durations, but on days with data, durations vary. This inconsistency
  might suggest fluctuating social interactions.
\end{itemize}

Best Guess if Experiencing Depression: Yes.

The data suggests variability in physical activity, time spent at home,
phone usage, and sleep patterns. This inconsistency could reflect
symptoms such as fatigue, loss of interest, and disturbed sleep,
aligning with some depression criteria. However, it's important to note
that this analysis is limited and cannot definitively diagnose
depression. A comprehensive assessment by a mental health professional
is necessary for an accurate diagnosis.

\begin{center}\rule{0.5\linewidth}{0.5pt}\end{center}

\section{Data Input Format}\label{appendix.data-input-format}
The following subsections include snippets detailing the input formats described in Section \ref{classification-methods}.
\subsection{CSV}
\begin{verbatim}
 date,total_distance_traveled(meters),time_at_home(minutes),location_entropy,...
 2019-04-29,49037.0,666.0,0.85,298.0,3.0,,,29.0,11430.0,40.0,1290.0,39.0,150.0,306.0,11.0
 2019-04-30,69171.0,555.0,0.87,274.0,4.0,16.0,,13.0,11427.0,49.0,1264.0,48.0,176.0,,
 …
\end{verbatim}
\subsection{Markdown}
\begin{verbatim}
 date|total_distance_traveled(meters)|time_at_home(minutes)|location_entropy|...
 2019-04-29|49037|666|0.85|298|3|nan|nan|29|11430|40|1290|39|150|306|11|
 2019-04-30|69171|555|0.87|274|4|16|nan|13|11427|49|1264|48|176|nan|nan|
 …
\end{verbatim}
\subsection{Tabular}
\begin{verbatim}
 date total_distance_traveled(meters) time_at_home(minutes) location_entropy …
 2019-04-29 49037.0 666.0 0.85 298.0 3.0 29.0 11430.0 …
 2019-04-30 69171.0 555.0 0.87 274.0 4.0 16.0 13.0 11427.0 …
 …
\end{verbatim}
\subsection{LaTeX}
\begin{verbatim}
 \begin{tabular}{lrrrrrrrrrrrrrrr}
 \toprule
 date & total_distance_traveled(meters) & time_at_home(minutes) & location_entropy & … \\
 \midrule
 2019-04-29 & 49037.0 & 666.0 & 0.850 & 298.0 & 3.0 & NaN & …\\
 2019-04-30 & 69171.0 & 555.0 & 0.870 & 274.0 & 4.0 & 16.0 & …\\
 \bottomrule
 \end{tabular}
\end{verbatim}

\section{Producing Reasoning Prompt}\label{appendix.reasoning-prompt}
Below is the prompt used to generate the reasoning samples used in Section \ref{producing-reasoning-samples}:

\begin{verbatim}
Role:  
You are a data analyst helping a psychiatrist understand human activity data. 

Task: 
You will be shown data gathered from a smartphone and smart watch worn by an individual. Your 
    goal is to analyze this data.
You are presented with the following: 
1. A table consisting of twenty-eight days of collected activity tracking data [Collected Data] 
2. Instructions on how to analyze the data [Instructions]

Collected Data 
date|total_distance_traveled(meters)|time_at_home(minutes)|...
|2019-05-06|11996|1012|...
…

Instructions 
Although the data does not contain explicit information on mood, trends in physiological 
    signals have been shown to correlate with mental health symptoms. Examine this data 
    and point out any specific trends or data points that could spark fruitful 
    conversation with a mental health professional.
\end{verbatim}

\section{Reasoning Grader Instructions}\label{appendix.rubric}
Thank you for taking the time to contribute to this study.

To start, please open this document that contains a table of data as
well as plots of the data.

Link to document: [LINK HERE]

You will now be asked to grade a series of 32 different statements
analyzing this data. Your goal is to check the accuracy of these
statements to ensure that references to the data are correct~

Here an explanation of the grading rubric. Please read this rubric
carefully:

\textbf{1. Does this response include numbers? (yes/no)}

\textbf{Yes}~-- at least some part of the response lists or quotes
specific numerical data or dates, regardless of correctness

\textbf{No}~-- the response does not include any specific numbers

~~~~~~~~~Note -- numbered lists don't count as numbers~

\textbf{2. Are these numbers consistent with the provided data?
(yes/no)}

\textbf{Yes}~-- all of the mentioned numbers or dates are included in
the provided data

\textbf{No}~-- some or all of the numbers or dates are not consistent
with the provided data, or there are no numbers (1 was answered ``No'')

For example:

\begin{itemize}
\item
  \begin{quote}
  The text statement says the highest sleep time occurred on May 9, but
  based on the graph you can see it is actually on June 2
  \end{quote}
\item
  \begin{quote}
  The text statement lists the lowest distance travelled as 127 meters,
  but the lowest distance traveled listed in the table is 1270
  meters~~~
  \end{quote}
\end{itemize}

\textbf{3. Does this response identify specific trends? (yes/no)}

\textbf{Yes}~-- the response makes statements relating to concepts like
minimum, maximum, averages, variability, upward or downward trends, etc.
as they pertain to the data

\textbf{No}~-- There is no statement of specific trends that relate to the included data.~

For example:

\begin{itemize}
\item
  \begin{quote}
  ``An increase in sleep might indicate a disturbance'' or "the
  individual makes phone calls" would not be a specific trend relating
  to the provided data
  \end{quote}
\item
  \begin{quote}
  ``The time spent asleep increased in the second half of the month''
  would be a specific trend relating to the data
  \end{quote}
\end{itemize}

\textbf{4. Are these trends consistent with the provided data? (yes/no)}

\textbf{Yes}~-- the listed trends are plausibly consistent with the
provided data table and/or plots

\textbf{No}~-- some or all of the listed trends are contradicted by the
provided data and/or plots or there are no specific trends (3 was
answered ``No'')

It is important to note that you should not evaluate further trends or reasoning as they may relate to, for example, mental health. For the purposes of grading these responses, it is only necessary to confirm if the response does or does not accurately describe the provided data. 
\newline
\newline
We anticipate it will take 1.5-2 minutes to grade each statement.

\end{document}